\documentclass{new_tlp}

\usepackage{amsmath}
\usepackage{graphicx}
\usepackage{subcaption}
\usepackage{multirow}

\usepackage{xurl}
\usepackage{verbatim}
\usepackage{amsfonts}
\usepackage{latexsym}
\usepackage{enumerate}
\usepackage{xcolor} 
\usepackage{graphicx} 
\usepackage{rotating}

\newcommand{\added}[1]{#1}

\newcommand{\nomeProblema}{\ac{NCD} Agenda}

\newcommand{\Horizon}{{\ensuremath{\mathbf H}}} 
\newcommand{\horizon}{{\ensuremath{h}}} 

\newcommand{\Servizi}{{\ensuremath{\mathbf S}}} 
\newcommand{\servizio}{{\ensuremath{s}}} 
\newcommand{\Srv}{{\ensuremath{S}}} 
\newcommand{\Pazienti}{{\ensuremath{\mathbf P}}} 
\newcommand{\paziente}{{\ensuremath{p}}} 
\newcommand{\Paziente}{{\ensuremath{P}}} 
\newcommand{\NumPazienti}{{{\ensuremath{N_P}}}}

\newcommand{\pacchetto}{{\ensuremath{\pi}}}

\newcommand{\Giorno}{{\ensuremath{{\mathtt{D}}}}}

\newcommand{\CUop}{{\ensuremath{Op}}} 
\newcommand{\Res}{{\ensuremath{CU}}} 

\newcommand{\prest}{{\ensuremath{{\mathtt{srvType}}}}} 
\newcommand{\service}{{\ensuremath{{\mathtt{sched\_service}}}}} 
\newcommand{\Prestazione}{{\ensuremath{S}}}
\newcommand{\capacity}{{\ensuremath{{\mathtt{shift}}}}} 
\newcommand{\schedule}{{\ensuremath{{\mathtt{schedule}}}}} 
\newcommand{\provides}{{\ensuremath{{\mathtt{provides}}}}} 
\newcommand{\precedes}{{\ensuremath{{\mathtt{precedes}}}}} 

\newcommand{\unfeasSubproblem}{{\ensuremath{{\mathtt{unfeas\_subproblem}}}}} 
\newcommand{\nogoodId}{{\ensuremath{{\mathtt{nogood\_id}}}}} 

\newcommand{\MinIntervalloNecessita}{{\ensuremath{\delta_{min}}}}
\newcommand{\MaxIntervalloNecessita}{{{\ensuremath{\delta_{max}}}}}

\newcommand{\OpWeakConstraint}{{\ensuremath{{\mathtt :\sim}}}}

\DeclareMathOperator*{\argmax}{arg\,max}

\usepackage{listings}

\usepackage{algorithm}
\usepackage{algorithmicx}
\usepackage{algpseudocode}
\algdef{SE}[DOWHILE]{Do}{doWhile}{\algorithmicdo}[1]{\algorithmicwhile\ #1}%

\usepackage{acronym}
\newacro{ASP}{Answer Set Programming}
\newacro{MP}{Master Problem}
\newacro{SP}{Sub Problem}
\newacro{IC}{Integrity Constraint}
\newacro{NCD}{Non-transmissible Chronic Disease}
\newacro{LBBD}{Logic-Based Benders Decomposition}
\newacro{NHS}{National Health Service}
\newacro{CP}{Clinical Pathway}
\newacro{MS}{Multi-Shot solving}

\newtheorem{example}{Example}

\begin{document}
\lstset{language=Prolog,
    basicstyle=\tt,         
    commentstyle=\textcolor[rgb]{0.00,0.59,0.00},       
    showstringspaces=false, 
    sensitive=true,	
    morekeywords = {disjunctive, cumulative, alldiffferent},
    deletekeywords={time, Op, op},
    mathescape=true         
 }

\title[Logic-Based Benders Decomposition in ASP]{Logic-Based Benders Decomposition in Answer Set Programming for Chronic Outpatients Scheduling}

\author[Cappanera, Gavanelli, Nonato and Roma]
{ Paola Cappanera \\ 
DINFO, Universit\`a degli Studi di Firenze, Italy \\ 
\email{paola.cappanera@unifi.it} \and 
Marco Gavanelli \\ 
DE, Universit\`a degli Studi di Ferrara, Italy \\
\email{marco.gavanelli@unife.it} \and 
Maddalena Nonato \\ 
DE, Universit\`a degli Studi di Ferrara, Italy \\
\email{maddalena.nonato@unife.it} \and 
Marco Roma \\ 
DINFO, Universit\`a degli Studi di Firenze, Italy \\
\email{marco.roma@unifi.it}}
\maketitle

\begin{abstract}
In Answer Set Programming (ASP), the user can define declaratively a problem and solve it with efficient solvers; 
 practical applications of ASP are countless  and
several constraint problems have been successfully solved with ASP.
On the other hand,  solution time usually grows  in a superlinear way (often, exponential) with respect to the size of the instance,
which is impractical for large instances.
A widely used approach is to split the optimization problem 
into sub-problems that are  solved in sequence,
some committing to the values assigned by others, and
reconstructing a valid assignment for the whole problem by juxtaposing the solutions of the
single sub-problems.
On the one hand this approach is much faster, due to the superlinear behavior;
on the other hand, it does not provide any guarantee of optimality:
committing to the assignment of one sub-problem can rule out the optimal solution from the search space.

In other research areas,  Logic-Based Benders Decomposition (LBBD) proved effective; 
in LBBD, the problem is decomposed into a Master Problem (MP) and one or several Sub-Problems (SP).
The solution of the MP is passed to the SPs, that can possibly fail. In case of failure, a no-good is returned
to the MP, that is solved again with the addition of the new constraint.
The solution process is iterated until a valid solution is obtained for all the sub-problems or  the MP is proven infeasible.
The obtained solution is provably optimal under very mild conditions.

In this paper, we apply for the first time LBBD to ASP, exploiting an application in health care
as case study.
Experimental results show the effectiveness of the approach.
We believe that the availability of LBBD can further increase the practical applicability of ASP technologies.
\end{abstract}


\section{Introduction}
\label{sec:intro}

\ac{ASP} is recently
 gaining momentum 
not only in the logic programming area but 
also in the constraint optimization and 
Operations Research communities.
ASP relies on the Stable Model Semantics \cite{StableModelsSemantics};
in an ASP formulation of a combinatorial problem, the solution is encoded in such a way that a stable model (or answer set) of the program corresponds to a solution of the problem.
Most ASP solvers divide the solution process into two steps:
a grounding phase, in which a ground program having the same stable models of the original program is generated, followed by a solving phase in which the
answer sets 
are computed.
%
The number of  applications is impressive, deserving five surveys in  a few years 
\cite{DBLP:journals/aim/ErdemGL16,DBLP:journals/ki/PaluDFP18,DBLP:journals/ki/FalknerFSTT18,10.1145/3191315,DBLP:journals/ki/Schuller18}.

However, many problems coming from real-life applications cannot be solved in reasonable time
 because their solution time is superlinear (often, exponential) with the instance size 
 and the instances are very large. 
As splitting large problems can simplify them,
the ASP literature reports many 
applications in which a difficult problem is 
split into two (or more) sub-problems that are then solved
in sequence (see \cite{guido2020scheduling,cardellini2021two,CarusoJLC23,DBLP:conf/padl/El-KholanySG22} just to name a few).
Such an approach could be described as follows.
Consider a problem $P$, in which a function $f(x,y)$ is maximized subject to a set of conditions $C(x,y), C_x(x), C_y(y)$, where $x$ and $y$ are two vectors of variables, that range respectively in the domains $D_x$ and $D_y$:
\begin{equation}
P: \qquad \max \{f(x,y) \mid x \in D_x, y \in D_y, C_x(x), C_y(y), C(x,y)\}.
\label{eq:problema_generico}
\end{equation}
Suppose that solving  $P$ is too computationally demanding.
The splitting approach would split $P$ into e.g., two sub-problems $P_x$ and $P_y$ (we simplify the exposition by considering  two sub-problems, although the approach could be extended to more levels and sub-problems), in which $P_x$ might be responsible of assigning values to $x$ variables, while $P_y$ to the $y$ variables.
One could then find the optimal assignment $x^*$ for $P_x$:
\begin{equation}
x^* = \argmax_x \{f(x,y) \mid x \in D_x, C_x(x)\}
\label{eq:soluzione_MP_generica}
\end{equation}
 then solve the remaining sub-problem
\begin{equation}
y^* = \argmax_y \{f(x^*,y) \mid y \in D_y, C_y(y), C(x^*,y)\}
\label{eq:soluzione_SP_generica}
\end{equation}
and finally provide the pair $(x^*,y^*)$ as proposed solution of the whole problem \eqref{eq:problema_generico}.

This approach can be considerably faster than solving the whole problem \eqref{eq:problema_generico}, due to the superlinear solving time;
on the other hand, it also has a number of issues.
First, the optimal solution $x^*$ for the first sub-problem might be impossible to extend to the $y$ variables, i.e., there might be no assignment to the $y$ variables such that $C(x^*,y)$ is satisfied.
In some applications one might be able to split the problem in such a way that for each value of the $x$ variables there always exists an assignment to the $y$ variables (and, indeed, in the aforementioned applications 
\cite{guido2020scheduling,cardellini2021two,CarusoJLC23,DBLP:conf/padl/El-KholanySG22}
the authors were able to find such an intelligent splitting); 
nevertheless, this limits the applicability of the splitting approach only to some specific applications.
Second, even if for $x^*$ there exists an assignment $y^*$ that satisfies all constraints $C(x^*,y^*)$, the pair $(x^*,y^*)$  might be not optimal for the global problem $P$.
In general, committing too early to the solution of one subproblem might prevent the optimal solution to be found.
This splitting approach could be thought of as a (very clever) greedy algorithm, in which one solves to optimality the first subproblem, greedily commits to it and then solves the second subproblem.

Benders Decomposition is a technique to decompose a problem into subproblems while retaining the ability 
to obtain the optimal solution and prove its optimality.
It was born in the realm of Operations Research, and relies on the duality theory of Linear Programming.
It was later extended  to  approaches that cannot rely on a duality theory in the so-called
\ac{LBBD} \cite{LBBD}.

In  \ac{LBBD}, a \ac{MP} is solved first, and provides tentative values that are passed to the sub-problems.
Consider the generic problem in Eq~\eqref{eq:problema_generico}; the optimal solution $x^*$ of the \ac{MP}
is obtained as in Eq.~\eqref{eq:soluzione_MP_generica}, and it is provided to the \ac{SP}.

Now, the sub-problem (Eq.~\eqref{eq:soluzione_SP_generica}) is solved using the suggested values $x^*$.
Two situations may occur: either the \ac{SP} is proven infeasible or its optimal solution $y^*$ is found.

In case of infeasibility, clearly the assignment $x^*$ is not acceptable for the whole problem (\ref{eq:problema_generico}), so a constraint 
that rules out $x^*$ 
is forwarded from  \ac{SP} to \ac{MP}.
Such a constraint is named a {\em feasibility cut} and it is added to the MP formulation.

In case \ac{SP} is feasible, its optimal solution $y^*$ is obtained together with the corresponding value of objective function $v^*=f(x^*,y^*)$;   again a new constraint, called {\em Benders cut}, is returned to the \ac{MP}.
Such constraint relates the \ac{MP} variables $x$ with  $v^*$
through a function $B_{x^*}(x)$,
imposing an upper bound on the value of the objective function:
\begin{equation}
B_{x^*}(x) \leq v^*
\label{eq:Benders_cut}
\end{equation}
How to formulate the function $B_{x^*}(x)$ is left to the designer of the \ac{LBBD} solution process, and it is a challenge, as
it can highly influence the efficiency of the decomposition.
In order to devise Benders cuts, the idea  is that 
the \ac{SP} solver proved that $v^*$ was optimal for the \ac{SP}, obtaining a proof of optimality 
that can be written as $(\forall y)\ f(x^*,y)\leq v^*$;
such a proof of optimality might also be extended to include values of $x$ different from $x^*$.
Delving into further details of this fascinating subject would distract us from the main topic of this paper; the interested reader can refer to
\cite{Hooker2019_LBBD_BookChapter} for an introduction to \ac{LBBD} and some examples of $B_{x^*}(x)$ functions in practical instances.

In both cases (SP feasible / infeasible), a new iteration is started: the \ac{MP} with the additional constraints is solved again, and the iteration continues until either the \ac{MP} is proven infeasible (and in such a case the whole problem is infeasible) or the optimal solution is found.
In any iteration, the optimal solution of the \ac{MP} is an upper bound of the whole problem \eqref{eq:problema_generico}:
since the \ac{MP} contains a subset of the constraints of the whole problem $P$,
it is actually one of its relaxations, so its optimal value is optimistic with respect to the real optimum of \eqref{eq:problema_generico}.
Again, in each iteration, the pair $(x^*,y^*)$, obtained by juxtaposing the optimal solution $y^*$ of the \ac{SP} \eqref{eq:soluzione_SP_generica} with the optimal solution $x^*$ of the \ac{MP}, is a valid solution, so its value $f(x^*,y^*)$ is a lower bound of the whole problem $P$.
If at any iteration the lower bound is equal to the upper bound, then $(x^*,y^*)$ is provably optimal for 
problem $P$.
A sufficient condition for termination of the iteration is that the bounds \eqref{eq:Benders_cut} are valid and the variables' domains   are finite \cite{LBBD}.

In various interesting cases, this procedure can be simplified; in particular if the sub-problems are feasibility problems, i.e.,
in Eq.~\eqref{eq:problema_generico} the objective function $f$ does not depend on the $y$ variables,
the \ac{LBBD} algorithm can be described as in Algorithm~\ref{alg:LBBD_feasibility}.

\begin{algorithm}
\caption{\label{alg:LBBD_feasibility} \ac{LBBD} scheme in case the sub problem is a feasibility problem.}
\begin{algorithmic}
\State $i\leftarrow 0$
\Repeat
	\State $i \leftarrow i+1$
	\State{$x_i^* = \Call{Solve}{MP}$}
    \State \textbf{if} MP is infeasible \textbf{then return} infeasible
	\State{$y_i^* = \Call{Solve}{SP,x_i^*}$}
    \State \textbf{if} {SP is infeasible}
		\textbf{then} generate a feasibility cut ruling out $x_i^*$ and add it to $MP$
\Until{SP is feasible}
\State \textbf{return} $(x_i^*,y_i^*)$
\end{algorithmic}
\end{algorithm}

In the rest of the paper, we will focus on decompositions in which the sub-problems are feasibility problems, i.e., the $y$ variables do not  explicitly occur in the objective function.

\added{LBBD can provide strong speedups in particular when there is a hierarchical relation between the solution of the master problem and that of the sub-problems, so that, once the assignment is found for the MP, the SP becomes easy in some sense (it could be a theoretically easy problem -- e.g., a problem in $P$ -- or  a problem that is experimentally found to be relatively easy). 
Further speedups can be obtained when once a solution for the MP is found, the rest of the problem $P$ consists of  independent sub-problems, that can be solved independently (even in parallel).}

In this work, we use for the first time \ac{LBBD} in an \ac{ASP}-based solving scheme.
We show its application on a challenging real-life problem coming from the healthcare domain.

\section{Case study}
\cite{ODS2021}, \cite{JLC23} addressed 
a scheduling problem involving chronic patients with comorbidities.
Many patients suffer from  so-called \acp{NCD},
such as diabetes, hypertension, cirrhosis, obesity, and so on. 
For most NCDs there exist well-assessed medical guidelines 
 involving periodic health services  to be delivered at  hospital premises --- 
think of  dialysis for patients with renal failure.
Most patients are not hospitalized but access  hospital premises 
as outpatients; 
many of them have more than one \ac{NCD} (comorbidity). 
Patients are assigned personalized care plans, i.e.,   \acp{CP}, 
that merge the  medical guidelines of all diagnosed NCDs,
customized to the specific patient.
A CP's health services are known a priori over a mid term horizon, which allows for  well in advance planning. 
%
%
Scheduling the health services of a CP means to assign a date, a time and an operator  to each  service the patient must receive at the hospital.
Such process can be challenging because 
appointment dates must comply as much as possible with the ideal frequency and
other time constraints, 
due to interference (a treatment may alter the result of an exam taken  after it) or  precedence (a consultancy requires recently taken exams).
Finally, if there is not enough availability within the public hospital,
a service can be provided by private health services at a higher cost for the National Health Service.
The centralized management of the CPs of all 
patients would
optimize the usage of public resources and ensure fairness.
This yields a very 
challenging problem that
we call  {\em \nomeProblema\ problem}.

\section{Related works} \label{sec:related}

Several  healthcare problems  have been  tackled 
with ASP  (see the review  \cite{alvrev2020}).
We recall the most notable contributions 
highlighting any decomposition. 

\cite{CarusoJLC23} schedule pre-operative exams for outpatients dividing
the problem  in two steps:
first, 
exam areas are staffed and  patients are given an appointment day;
second,  exams starting times are set, complying with first level decisions, maximizing the served patients  and minimizing  waiting time.
Each phase is executed  once, with no feedback; to ensure feasibility in phase two, demand is overestimated in phase one. 

\cite{guido2020scheduling} schedule multi appointments for rheumatic outpatients at a Hospital Day Service. 
Patients are partitioned into three classes with decreasing priority; to reduce computing time the schedule is computed separately for each priority class.

\cite{alviano2018nurse} address the nurse (re)scheduling problem,  improving on the representation of hospital and work balance constraints by \cite{dodaro2017nurse}.
Nurse scheduling consists of determining a shift assignment for each nurse for a given planning horizon such that working hours, shift mix, and rest days comply with  hospital rules.
Rescheduling is due in case of nurse temporary absences, and consists of feasibly scheduling  vacant duties, while minimizing deviations from the original schedule. 

Appointment scheduling 
for chemotherapy treatments (\cite{dodaro2021TPLPchemio})
must deal with the availability of special equipment, 
that is assigned to a patient for the whole session.
A treatment encompasses up to 4 subsequent steps, some of which are optional, whose duration is patient dependent and known.
In case of multiple treatments, a treatment frequency is given. 
A weekly problem is solved, as well as a rescheduling one. 

Rehabilitation sessions for inpatients are scheduled by \cite{cardellini2021two}. 
Two types of resources are present: gyms and operators.
Solution quality criteria and constraints include:   continuity of care and  preferred time slots on the patient side, and  workload balancing and abiding by  working rules on the operator side.
The daily problem is decomposed into two subsequent decision phases. In the first, the board, patients are assigned to operators;
in the second, the agenda, a starting and ending time is set for each session according to the board.
As there is no feedback, there is no guarantee that a feasible board-compatible agenda exists.
To this aim, potential overlapping are admitted: some sessions are partially
turned from one-to-one care to supervision (one operator supervises a few patients at a time).

Finally, ASP has been proved effective in 
the (re)scheduling of operating rooms (OR).
A planned surgery requires a free bed  at the specialty ward or at intensive care units, starting from surgery date for the predicted length of stay 
(\cite{dodaro2021operating}), and a bed at the post anaesthetic care unit for  temporary post-surgery staying (\cite{galata2021asp}). 
Since a surgical team is made of  surgeons, anesthesiologists, and nurses,
the whole surgery slot must be fully contained into the current working shift of each team member.
Based on its specialty, priority, special needs, and expected duration, a request is assigned a day and a time during the OR time blocks 
reserved to its specialty. 

Out of the healthcare domain, \cite{DBLP:conf/padl/El-KholanySG22} present  a  decomposition scheme in ASP for Job Shop Scheduling, driven by a machine learning algorithm.
There is no feedback from the \acp{SP} to the \ac{MP}, each sub-problem is solved
only once, and thus  the resulting algorithm cannot prove optimality of the found solution (it is a heuristic algorithm). The approach is further improved in \cite{DBLP:journals/tplp/El-KholanyGS22}.

In conclusion, we observe 
that ASP proved to be able to capture and easily represent the complex features of several challenging problems.
Decomposition schemes are often implemented, motivated by the need for solving large instances in a reasonable time,
however they are implemented in such a way that
the optimal solution could be overlooked and optimality cannot be guaranteed.
To the best of our knowledge, we are the first to propose the use of \ac{LBBD} in ASP.
Out of the ASP area, LBBD has been successfully applied in Integer Linear Programming \cite{Mannino2016} and Constraint Programming frameworks, often as an hybrid algorithm. \cite{DBLP:conf/cp/Fazel-ZarandiB09} use a hybrid CP-MILP approach for facility location, \cite{Milano2011} for resource allocation for multicore platforms, just to name a few.
Recently, in \cite{Zhu2023} two major problems in manufacturing - usually solved in pipe, to the detriment of optimality - were handled together exploiting LBBD and a clever CP based formulation.
A comprehensive survey can be found in \cite{Hooker2019_LBBD_BookChapter}.
LBBD was also applied to a railway timetabling problem using Satisfiability Modulo Theory \cite{DBLP:journals/eor/LeutwilerC22}.

Finally, the splitting set theorem  \cite{SplittingSetThm} provides syntactic conditions under which the stable models of a program can be obtained extending the stable models of one subprogram. LBBD is applied on a different level: the level of modelling an optimization problem and decomposing it into subproblems, even if the splitting set theorem could be exploited to have synergies with LBBD.
\section{\nomeProblema\ formalization}\label{sec:problem}

Let us consider 
a planning horizon (set of available days) $\Horizon = \{ 1,\dots,\horizon \}$, 
a set of health services $\Servizi$,  
and a set of patients $\Pazienti = \{ \paziente_1, \dots, \paziente_\NumPazienti \}$.
For each patient $\paziente$,  a \ac{CP} is known, 
consisting of a set of \emph{packets}. 
Each packet \pacchetto\ is a set of services to be delivered on the same date, even if they are provided by different care units.

The appointment dates of each CP should satisfy the following  \emph{CP constraints}:

{\em Frequency}: 
Often a pathway contains sets of packets corresponding to recurring services; for each packet there is an ideal date (ensuring that the patient is serviced with the correct frequency) and the packet should be scheduled within a tolerance from the ideal date.
The tolerance  depends on the pathway and it is such  that the tolerance windows of consecutive occurrences of the same packet are disjoint.

{\em Interdiction}: if $s_i$ 
interdicts service $s_j$ for $\delta$ days and $s_i$ is scheduled in $\tau(s_i)$, then $\tau(s_j)$,
the appointment date of $s_j$, is such that $\tau(s_j)\notin [\tau(s_i),\tau(s_i)+\delta]$.
Interdiction constraints are always satisfied if one of the two services is not scheduled.

\emph{Necessity}: if $s_i$ requires $s_j$, an interval $[\MinIntervalloNecessita,\MaxIntervalloNecessita]$ is provided; service $s_j$ should be scheduled on day $\tau(s_j) \in [\tau(s_i)+\MinIntervalloNecessita,\tau(s_i)+\MaxIntervalloNecessita]$
and cannot be scheduled in the (right-open) interval $[\tau(s_i),\tau(s_i)+\MinIntervalloNecessita)$.

A second class of constraints 
concerns resource assignment:  each service $\servizio$ has a service type 
and a duration; 
each scheduled service should be assigned to an operator of the care unit 
that provides that service type. 
Each operator at the care unit has a working shift (start and end time, potentially empty) for each day in the horizon.
The following \emph{daily agendas constraints} hold: 
$i$) all services provided by an operator should be completed within the operator shift and  
$ii$) without overlaps \emph{(no patient overlapping)},
$iii$) each patient cannot receive two services in parallel \emph{(no service overlapping)}, 
$iv$) a service cannot be interrupted and resumed at a later time \emph{(no preemption)},
as well as 
$v$) a scheduled service is delivered by a single operator \emph{(no split-service)}.

A feasible schedule assigns  an appointment date $\tau(\pacchetto)$
to each scheduled packet \pacchetto,
as well as a time and an operator to the services of \pacchetto,
so that all constraints are satisfied.
If a packet is not scheduled, the patient will receive the same services from a private clinic at a higher cost. 
The objective is to maximize the number of  scheduled packets. 

For example, in Figure~\ref{fig:SchemaInterazione} the set of patients is $\Pazienti = \{p1,p2\}$; the pathway of  $p1$ is made of just one packet $\pacchetto_1$ which includes two services; the color (red or blue) represents the service type  and each service is associated with the care unit of the same color. Care unit 1 (red) has 4 time slots of availability on day 1, 3 slots on day 2, and 2 slots on day 3. Note also the different start times of the operators' shifts: the operator of care unit 1 on day 3 starts earlier than that of care unit 2.
\section{ASP approaches}
\label{sec:ASP}

\ac{ASP} 
relies on the Stable Model Semantics \cite{StableModelsSemantics}.
An ASP program is a set of clauses $\mathtt{h \mbox{:-} b_1,\dots,b_n}$, where $\mathtt{h}$ is an atom $\mathtt{p(t_1,\dots,t_m)}$ or a choice $\mathtt{\{p(t_1,\dots,t_m)\}}$ and $\mathtt{b_i}$ can either be literals of the form $\mathtt{[not] p(t_1,\dots,t_k)}$, possibly followed by a condition
$\mathtt{:c_1,\dots,c_k}$, or an aggregate
\verb|#sum|$\mathtt{\{t_1,\dots,t_m:c_1,\dots,c_k\} \circ n}$ where $\circ$ is a comparison operator $<,=,>=,\dots$.
A clause without head is called an \ac{IC}, and its body must evaluate to false in every Stable Model (or Answer Set) of the program.
Optimization components can be added by means of 
weak integrity constraints, with syntax $\OpWeakConstraint body$;
the aim will be to find an answer set that satisfies all \acp{IC} while satisfying the maximum number of weak \acp{IC}.
For the full ASP syntax see \cite{ASPCore2}.

We recap the ASP formalization of the agenda component of the \nomeProblema\  \cite{ODS2021} in Sect~\ref{sec:masterMonolitico}, 
and the scheduling of services within the day 
 in Sect~\ref{sec:slaveMonolitico}.
 The \ac{LBBD} approach is developed in Section~\ref{subsec:LBBD1}.




\subsection{Scheduling services - date assignment}
\label{sec:masterMonolitico}

The input data is provided by the following predicates:
\begin{itemize}
    \item {\tt occurrence\_to\_schedule(Patient,Packet)} provides the packets that should ideally be scheduled for each patient;
    the ideal date is {\tt ideal\_date(Patient, Packet, IdealDate)}, but a tolerance is accepted; predicate
    {\tt within\_tol(Packet, Day, IdealDate)} checks if {\tt Day} stands within the tolerance.
    \item The set of available days for the scheduling is provided by {\tt day(D)};
    \item {\tt service\_in\_packet(Srv,Pck)} means that service {\tt Srv} belongs to packet {\tt Pck}
    \item {\tt necessity(Service1,Service2,(Dmin,Dmax))} means that if {\tt Service1} is scheduled on day $d_1$, {\tt Service2} should be scheduled in the interval $[d_1+\mathtt{Dmin},d_1+\mathtt{Dmax}]$.
    \item {\tt interdiction(Service1,Service2,Ndays)} states that {\tt Service2} cannot be scheduled for {\tt Ndays} after {\tt Service1}.

\end{itemize}

The ASP program for the scheduling of packets to the available days (Listing~\ref{listing:MP}) follows the classical generate \& test methodology.
The generation part (lines \ref{line:MP_genStart}--\ref{line:MP_genEnd}) tries to assign a date {\tt Day} to each 
{\tt Packet} within the given tolerance from the ideal date.

\begin{lstlisting}[basicstyle=\ttfamily,float,caption=Date assignment,label=listing:MP,numbers=left, numberstyle=\tiny,frame=tb,escapeinside={(*@}{@*)},numberblanklines=false]
0{schedule(Pat,Pck,Day):day(Day), within_tol(Pck,Day,Ideal)}1(*@ \label{line:MP_genStart} @*)
:- occurrence_to_schedule(Pat,Pck), 
  ideal_date(Pat,Pck,Ideal).
 
$\service$(Pat, Service, Day) :- $\schedule$(Pat, Packet, Day),(*@ \label{line:MP_sched_service} @*)
  service_in_packet(Service,Packet). (*@ \label{line:MP_genEnd} @*)
$\OpWeakConstraint$ occurrence_to_schedule(Pat,Pck),(*@ \label{line:MPweak} @*) not schedule(Pat,Pck,_).

:- $\service$(Pat,Srv1,Day1), (*@ \label{line:MP_IC_interdiction} @*)$\service$(Pat,Srv2,Day2),
   interdiction(Srv1,Srv2,Ndays), 
   Day2 >= Day1, Day2 <= Day1 + Ndays.
:- $\service$(Patient,Srv1,Day1), (*@ \label{line:MP_IC_necessity} @*)necessity(Srv1,Srv2,_),
   not satisfied_necessity(Patient, Srv1, Srv2).
   
satisfied_necessity(Pat, Srv1, Srv2) :- 
  $\service$(Pat,Srv1,Day1), $\service$(Pat,Srv2,Day2), 
  necessity(Srv1, Srv2, (Dmin,Dmax)), 
  Day2 >= Day1 + Dmin, Day2 <= Day1 + Dmax.
satisfied_necessity(Pat,Srv1,Srv2) :- 
  $\service$(Pat, Srv1, Day1),
  necessity(Srv1, Srv2, (Dmin,Dmax)), Day1 + Dmax > horizon.
  
:- $\service$(Pat,Srv1,Day1), (*@ \label{line:MP_IC_necessity_excluded} @*)$\service$(Pat,Srv2,Day2),
  necessity(Srv1,Srv2,(Dmin,_)), Day1<Day2, Day2<=Day1+Dmin.
\end{lstlisting}

As a packet could be not scheduled at all, the number of packets scheduled within the horizon will be maximized by the weak constraint in line~\ref{line:MPweak}.
%
%
%
The \ac{IC} in line~\ref{line:MP_IC_interdiction} deals with interdiction constraints:
{\tt Srv1} and {\tt Srv2} are two services  for the same patient and  the first interdicts the second for {\tt Ndays}.
%
The IC in line~\ref{line:MP_IC_necessity} ensures that each necessity constraint is satisfied. Predicate {\tt satisfied\_necessity} declares that the necessity must be either satisfied within the horizon, or assumed to be satisfied beyond it,
while the previous condition $\tau(s_j)\notin [\tau(s_i),\tau(s_i)+\delta_{min})$ is dealt with by the  \ac{IC} in line~\ref{line:MP_IC_necessity_excluded}.






\subsection{Daily agendas}
\label{sec:slaveMonolitico}

In the \ac{ASP} formalization 
in Section~\ref{sec:masterMonolitico},
services are assigned a date, but daily agendas are not handled, i.e., neither a starting time is given, nor services are assigned to specific operators at the various care units.  
The program for the daily agendas, reported in Listing~\ref{listing:SP}, uses the following input predicates, in addition to those in Section~\ref{sec:masterMonolitico}:
\begin{itemize}
    \item $\prest{\mathtt{(s,cu,dur)}}$: the list of services, together with the care unit {\tt cu} that can provide it and the duration {\tt dur};    
    \item $\capacity{\mathtt{(D, \Res, \CUop, St, Dur)}}$:  
    for each operator ${\mathtt{\CUop}}$,  the start time ${\mathtt{St}}$ and the duration ${\mathtt{Dur}}$ of the shift are provided in each day $\mathtt{D}$, together with the care unit ${\mathtt{\Res}}$ the operator works in.
    
\end{itemize}

\begin{lstlisting}[basicstyle=\ttfamily,float,caption=Daily agendas,label=listing:SP,numbers=left, numberstyle=\tiny,frame=tb,escapeinside={(*@}{@*)},numberblanklines=false]
1{start_time$(\mathtt{Pat, \Srv, Day, Start})$ : time(Start)}1 :- 
   $\service(\mathtt{Pat,Day, \Srv})$. (*@ \label{line:SP_start_time} @*)
1{$\provides(\mathtt{\CUop,Pat,\Srv,Day})$ : $\capacity(\mathtt{Day,\Res,\CUop,}$_$,$_$), \prest(\mathtt{\Srv,\Res,}$_$)$}1 :- (*@ \label{line:SP_operator} @*)
   $\service(\mathtt{Pat, \Srv, Day})$.

$\precedes$(P1,S1,P2,S2,Day) :- $\prest$(S1,_,D1), 
   Start1 + D1 <= Start2, (*@ \label{line:SP_precedes} @*)
   start_time(P1,S1,Day,Start1), 
   start_time(P2,S2,Day,Start2).

:- not $\precedes$(P,S1,P,S2,Day), not $\precedes$(P,S2,P,S1,Day), (*@ \label{line:SP_ic_no_overlap1} @*)
   $\service$(P, S1, Day), $\service$(P, S2, Day), 
   S1 != S2.
:- not $\precedes$(P1,S1,P2,S2,Day), 
   not $\precedes$(P2,S2,P1,S1,Day), (*@ \label{line:SP_ic_no_overlap2} @*)
   provides(CU,Op,P1,S1,Day), provides(CU,Op,P2,S2,Day),
   $\service$(P1,S1,Day), $\service$(P2,S2,Day), P1!=P2.

:- $\service$(P,S,Day), $\prest$(S,CU,Dur), (*@ \label{line:SP_ic_within_shift1} @*)
   $\provides$(Op,P,S,Day), start_time(P,S,Day,Start), 
   Start + Dur > StartShift + DurShift, 
   $\capacity$(Day,CU,Op,StartShift,DurShift).
:- $\service$(P,S,Day), $\provides$(Op,P,S,Day), (*@ \label{line:SP_ic_within_shift2} @*)
   Start < StartShift, start_time(P,S,Day,Start), 
   $\capacity$(Day,CU,Op,StartShift,DurShift).
\end{lstlisting}

In the generate part of the daily agenda program (Listing~\ref{listing:SP}), each scheduled service is assigned a start time (line~\ref{line:SP_start_time}) and an operator of the care unit that provides the required service type (line~\ref{line:SP_operator}), leveraging on the daily assignment defined by predicate \service\  (line~\ref{line:MP_sched_service} of Listing~\ref{listing:MP}).


In order to avoid overlapping between services of the same patient (constraint $iii$ of the daily agenda problem) or delivered by the same operator (see $ii$), we define  predicate \precedes\ (line~\ref{line:SP_precedes}), stating that service {\tt S1} of patient {\tt P1} precedes {\tt S2} of {\tt P2} if they are scheduled on the same day and {\tt S1} terminates before or at the same time as  {\tt S2}  starts.
Now 
if two services {\tt S1} and {\tt S2} are for the same patient or are
provided by the same care unit operator $\mathtt{\CUop}$ in the same {\tt Day}, one of the two services must precede the other (\acp{IC} in lines \ref{line:SP_ic_no_overlap1} and \ref{line:SP_ic_no_overlap2}).
%
%
%
%
Finally, ICs in lines \ref{line:SP_ic_within_shift1} and \ref{line:SP_ic_within_shift2} state that each service should be scheduled within the working shift of the operator who  delivers it (constraint $i$ of the daily agenda).
   

\subsection{LBB decomposition} \label{subsec:LBBD1}
The \ac{ASP} formalization of 
Sections~\ref{sec:masterMonolitico} and \ref{sec:slaveMonolitico} correctly solves the \nomeProblema\ problem; 
on the other hand solving such a difficult problem in a monolithic approach does not scale well with the size of the instance.
To speedup the solving process while retaining the completeness of the search,
we apply \ac{LBBD} (\cite{LBBD}).


The \nomeProblema\ \added{can be cast as in Eq~\ref{eq:problema_generico} in which the} \ac{MP}  (based on the ASP program in Section~\ref{sec:masterMonolitico}) maximizes the number of scheduled packets while assigning a day to each scheduled packet; \added{while} a series of \acp{SP} (based on the program in Section~\ref{sec:slaveMonolitico}), one for each day, assign a time and 
an operator
to each service.
Decomposing the program  this way provides a strong improvement, 
since the \acp{SP} are independent problems, one per day, and they could even be solved in parallel (although in our experiments we do not exploit such parallelism in order to have a fair comparison with the monolithic approach). 
On the other hand, some \acp{SP} could be infeasible, since the master problem 
does not contain all constraints of  the \nomeProblema.

The \ac{MP} is an optimization problem, while each \ac{SP} is a satisfiability problem. 
In case all \acp{SP} are satisfiable, the optimal solution of the \ac{MP} is also the optimum of the whole \nomeProblema\ problem.
Otherwise, if one of the \acp{SP} is infeasible, a no-good is returned to the \ac{MP} conveying the information that the particular set of health services the \ac{MP} has assigned to that day cannot be feasibly served. 
Then, the \ac{MP} is solved again, with the additional no-good, which avoids looping.
Convergence occurs when each \ac{SP} admits a feasible solution. Such solution is 
 provably optimal.

In particular, the unfeasible \ac{SP} returns to the \ac{MP} the set of packets that could not be scheduled on that day, as a set of facts of the form
\begin{lstlisting}
unfeas_subproblem(patient,packet,day,gid)
\end{lstlisting}
together with a fact \lstinline|nogood_id(gid)|,
where {\tt gid} is a unique identifier for the group of packets.
A new version of the \ac{MP} is then generated, appending to the previous code the new facts \lstinline|unfeas_subproblem| and \lstinline|nogood_id|, and together with
the following \ac{IC}, that avoids generating schedules for the same day including all the packets in the no-good
\begin{equation}
\begin{array}{l}
{\mathtt{
\mbox{:-} \schedule(Pat,Pck,Day): \unfeasSubproblem(Pat,Pck,Day,Gid);
 \nogoodId(Gid).}}
\end{array}
\label{eq:MP_accetta_nogood}
\end{equation}

\begin{figure}
    \centering
    \includegraphics[width=\linewidth]{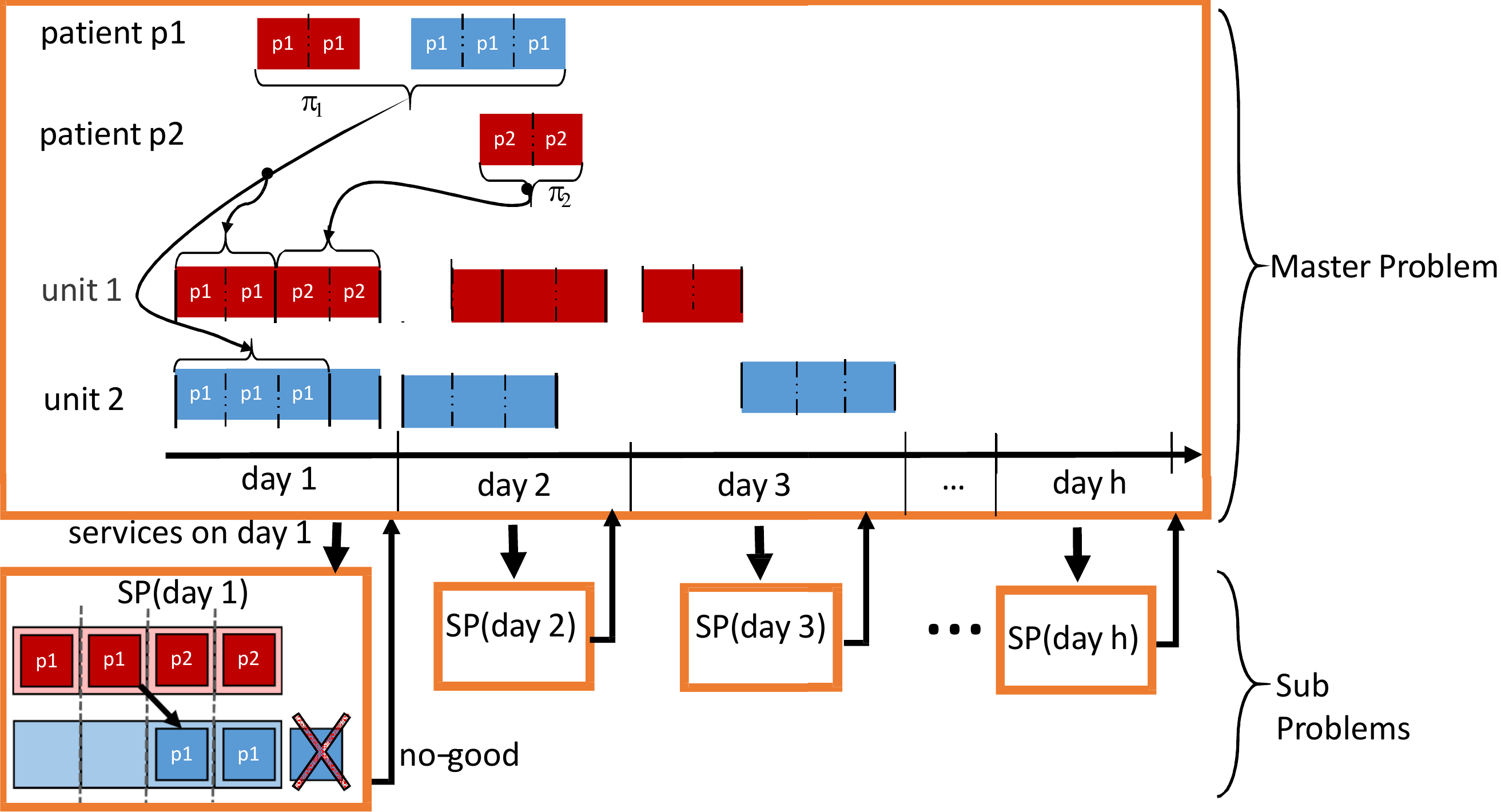}
    \caption{Interaction scheme in LBBD}
    \label{fig:SchemaInterazione}
\end{figure}

\begin{example}
\label{ex:interazioneMP_SP}
Consider the example in Figure~\ref{fig:SchemaInterazione}, already introduced in Sec \ref{sec:problem}.
The master problem may schedule both patients on day 1, as depicted.
In such a case, the \ac{SP} for day 1 will detect infeasibility and  return a no-good  to the \ac{MP}, stating that both packets cannot fit together on that day:
\begin{equation}
\begin{array}{ll}
\mathtt{\unfeasSubproblem(p1,pck1,day1,gid1).}\\
\mathtt{\unfeasSubproblem(p2,pck2,day1,gid1).}  &    \mathtt{\nogoodId(gid1).}
\end{array}
\label{eq:esempioNogood}
\end{equation}
The \ac{MP} receives the no-good and, in the following iterations, it will contain the \ac{IC} in Eq.~\eqref{eq:MP_accetta_nogood}, which is grounded into
$$
{\mathtt{\mbox{:-} \schedule(p1,pck1,day1), \schedule (p2,pck2,day1).}}
$$
so that, in the following iterations, at most one of the two packets can be assigned to day 1.
Since a limited number of options are present, after a certain number of iterations yielding similar no-goods,  the \ac{MP} will schedule $p_1$ on day 3 and $p_2$ on any other day. 
The process then stops and returns a feasible solution.
\end{example}

In \ac{LBBD}, in  order to speedup convergence, it is worth
strengthening the \ac{MP} by adding a relaxed version of some of the \acp{SP} constraints.
 In our case,  in the \ac{MP} we  avoid any reference to the timing within the day, but we state that, for each day  and  each care unit,
the total duration of the services assigned to that care unit cannot exceed
the sum of the shift duration 
of all the operators of that unit:
\begin{equation}
\begin{array}{l}
\mathtt{\mbox{:-} day(\Giorno), total\_time(\Giorno, \Res, TotTime),}
\\
\mathtt{\#sum\{Dur,\Paziente,\Prestazione : \service(\Paziente, S, \Giorno), \prest(\Prestazione, \Res, Dur)\}>TotTime.}
\end{array}
\label{eq:icSommaDurate}
\end{equation}
\begin{lstlisting}
total_time($\mathtt{\Giorno}$, $\mathtt{\Res}$, Time):- $\prest$(_,$\mathtt{\Res}$,_), day($\mathtt{\Giorno}$),
    #sum{Dur,$\Giorno$,$\mathtt{\Res}$:$\mathtt{\capacity(\Giorno, \Res, Op,}$ _, Dur$)$} = Time.
\end{lstlisting}
Note that predicate {\tt total\_time} is grounded into a set of facts by the grounder {\tt gringo}.

The \ac{MP} solution may violate the daily agenda constraints: 
 as shown in Example~\ref{ex:interazioneMP_SP}, the \ac{MP} may schedule both packets on day 1 or on day 2.
However, 
with the availabilities depicted in Fig.~\ref{fig:SchemaInterazione}, \ac{IC}  (\ref{eq:icSommaDurate})  forbids the \ac{MP} to schedule both packets on day 3, as the red care unit provides only two time units.

\subsection{Multi-shot solving}\label{sub:multishot}

The interaction scheme in Section~\ref{subsec:LBBD1} can be implemented by a script that iteratively invokes the ASP solver on the \ac{MP}, on the \acp{SP} and adds to the \ac{MP} code the no-goods generated by the \acp{SP}.
However, in this way the \ac{MP} should be solved from scratch at every iteration,  losing information about the clauses learnt in the previous iteration.


Recent versions of Clingo \cite{ClingoMultiShot} allow the customization of ASP solving processes that deal with continuously changing logic programs, called {\em multi-shot solving}.
The solving process can be controlled through commands written in Python. 
It is possible to integrate non-ground input rules into subprograms having a name and a list of parameters, and that are introduced by the  \ensuremath{{\mathtt{\#program}}} directive.
A dedicated subprogram \ensuremath{{\mathtt{base}}}  gathers all the rules that are not included in a
subprogram \cite{kaminski2017tutorial}.
By default, Clingo grounds and solves just the \ensuremath{{\mathtt{base}}} program, but we can add control in Python using a main routine taking as argument a control object representing the state of Clingo.

In order to exploit Multi-Shot solving in the devised \ac{LBBD}, the following subprogram
is added to the ASP formulation of the \ac{MP}:
\begin{lstlisting}
#program nogood(pat,pck,day,gid).
unfeas_subproblem(pat,pck,day,gid). nogood_id(gid).
:- schedule(Pat,Pck,D) : unfeas_subproblem(Pat,Pck,D,Gid); 
    D=day, Gid=gid.
\end{lstlisting}
The parameters {\tt pat}, {\tt pck} and {\tt day} correspond to the arguments of the \ensuremath{{\mathtt{schedule}}} predicate, and {\tt gid} identifies a group of schedules that produced an inconsistent \ac{SP},
as in Eq.~\eqref{eq:MP_accetta_nogood}.

The \ensuremath{{\mathtt{nogood}}} program can be grounded incrementally from a Python script, passing a list $L$ of ground terms; each term contains 
the 4 parameters of \ensuremath{{\mathtt{nogood}}}.
In the \ac{LBBD}, the parameters 
will be the
schedules that made unfeasible one of the \acp{SP},
provided as a 
 set of ground facts \ensuremath{{\mathtt{unfeas\_subproblem}}}.

Algorithm~\ref{alg:LBB_MultiShot} gives the pseudocode that 
 controls the solving process, solving in sequence the  MP ({\em``base"}), then each \acp{SP} with an external call to Clingo for each day. 
Afterwards, from failing \acp{SP} 
the group of packets scheduled by the MP in the specific day is added as a no-good.
This information is added to the \ac{MP} grounding the \ensuremath{{\mathtt{nogood}}} program, and the process loops until no new constraints are added, i.e., all \acp{SP} are satisfiable.

\newcommand{\VariabileAnswerSet}[1]{{\ensuremath{AS^{#1}}}}

\begin{algorithm}
\caption{\label{alg:LBB_MultiShot} LBBD with multi-shot solving}
\begin{algorithmic}

\State \textbf{function} {LBBD}($prg$)
\State {prg.ground($[("base", [])]$)}  \Comment{ground the base program (MP)}
\Do
\State \VariabileAnswerSet{MP} = {prg.solve()   \Comment{solve the ground program(s)}}
\State \textbf{if} {MP has no solution} \textbf{then return} Unsatisfiable
\State NGs = $\emptyset$
\For{$day$ in $horizon$}
\State \VariabileAnswerSet{SP_{day}}=\Call{solve\_SP}{$day, \VariabileAnswerSet{MP}$}   \Comment{solve a Subproblem for each day}
\State \textbf{if} {$SP_{day}$ has no solution}
\textbf{then} \\ 
    \qquad \qquad NGs = NGs $\cup$ \Call{compute\_nogoods}{$day, \VariabileAnswerSet{MP}$}
\EndFor
 \State prg.ground($[``nogood", NGs]$)   \Comment{ground nogood program with parameters}
\doWhile{$NGs \neq \emptyset$}
\State \Return $\VariabileAnswerSet{MP} \cup \bigcup_{day} \VariabileAnswerSet{SP_{day}}$
\end{algorithmic}
\end{algorithm}

\section{Experimental results}
\label{sec:experiments}
We implemented an instance generator based on well assessed and publicly available medical guidelines for the most common NCDs\footnote{\url{https://salute.regione.emilia-romagna.it/cure-primarie/diabete/gestione-integrata-del-diabete-mellito-di-tipo-2-2017/}, \\
\url{https://www.regione.toscana.it/documents/10180/23793180/ALL+A+23-2019+PDTA-Diabete.pdf/f1e8ea87-145f-08c4-6c3d-16b69f5f43c2?t=1578658143393/}}.
The generator allocates resources on a weekly basis and replicates the allocation in each week of the given time horizon, then it generates CPs.
Specifically, we have: (i) 5 CUs, each of them with a daily capacity (expressed in number of slots) drawn with uniform probability in the range [24, 60]; (ii) a number of operators drawn with uniform probability in [1,4] on each day of the week for each CU; (iii) services with a duration  (expressed in number of slots) in [6,15] and associated with a CU; 
(iv) packets made of  4 services at most;
(v) a given number of  patients with a number of CPs in [1,4]. The probability of assigning a number of CPs to a given patient is inversely proportional to the number of CPs itself. 

 For each number of patients in $\{10,20,40\}$ and length of the planning horizon in \{30,60\} days, 20 instances are generated, summing up to 120 instances. 
 Experiments were run with Clingo 5.6.2 with a time limit of one hour on a Ubuntu 22.04.1 LTS  OS, Intel(R) Xeon(R) CPU E5-2430 v2 @ 2.50GHz machine with a 32GB System Memory.

\begin{figure}
    \centering
    \includegraphics[width=\textwidth]{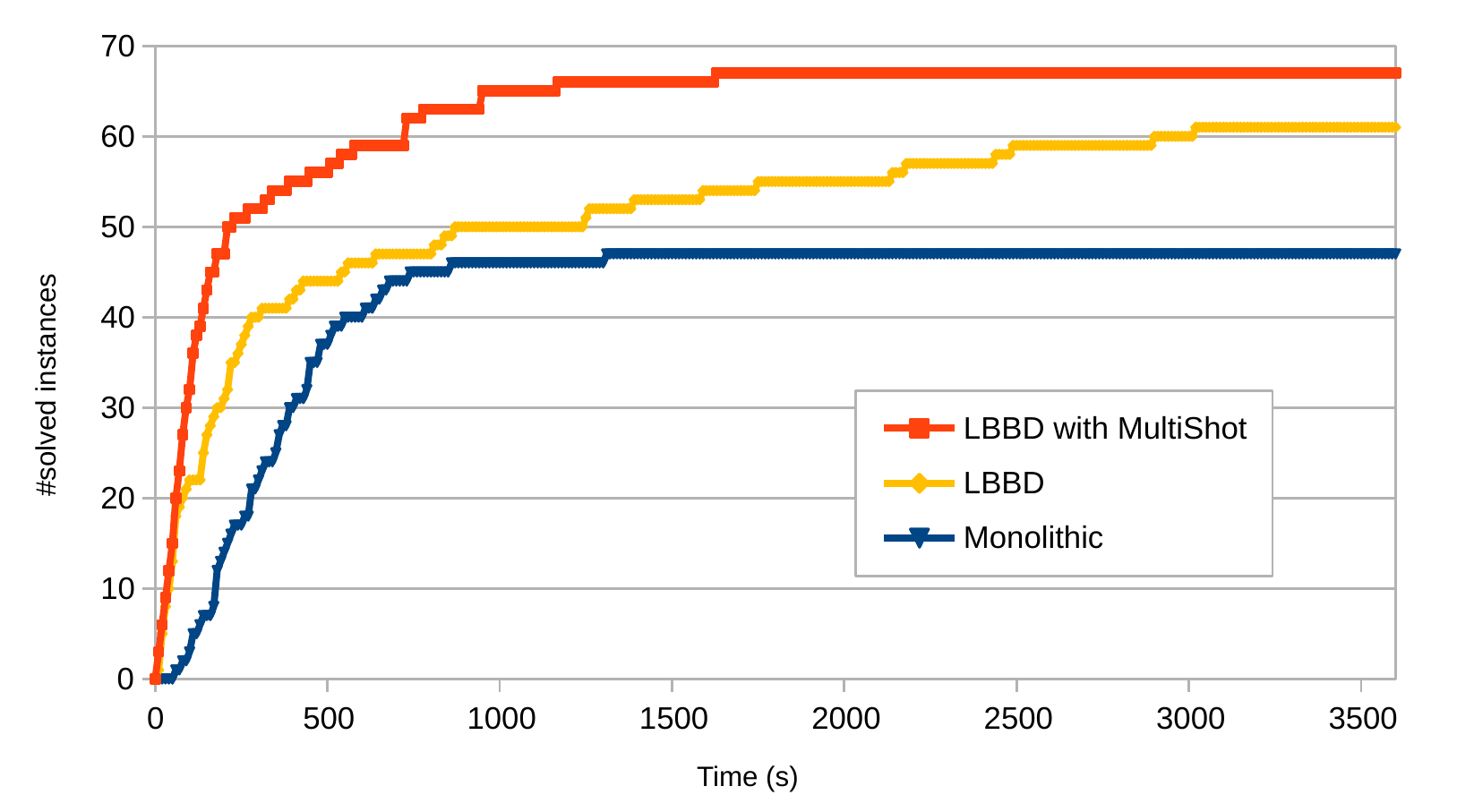}
    \caption{\label{fig:CactusPlot} Number of solved instances vs running time}
\end{figure}

 Figure \ref{fig:CactusPlot} shows on the $y$-axis the number of instances solved by each of the three approaches within a given computation time reported on the $x$-axis. The decomposition approach lets one solve to optimality between 30\% and 42\% more instances.

\begin{figure}
    \centering
    \includegraphics[width=\textwidth]{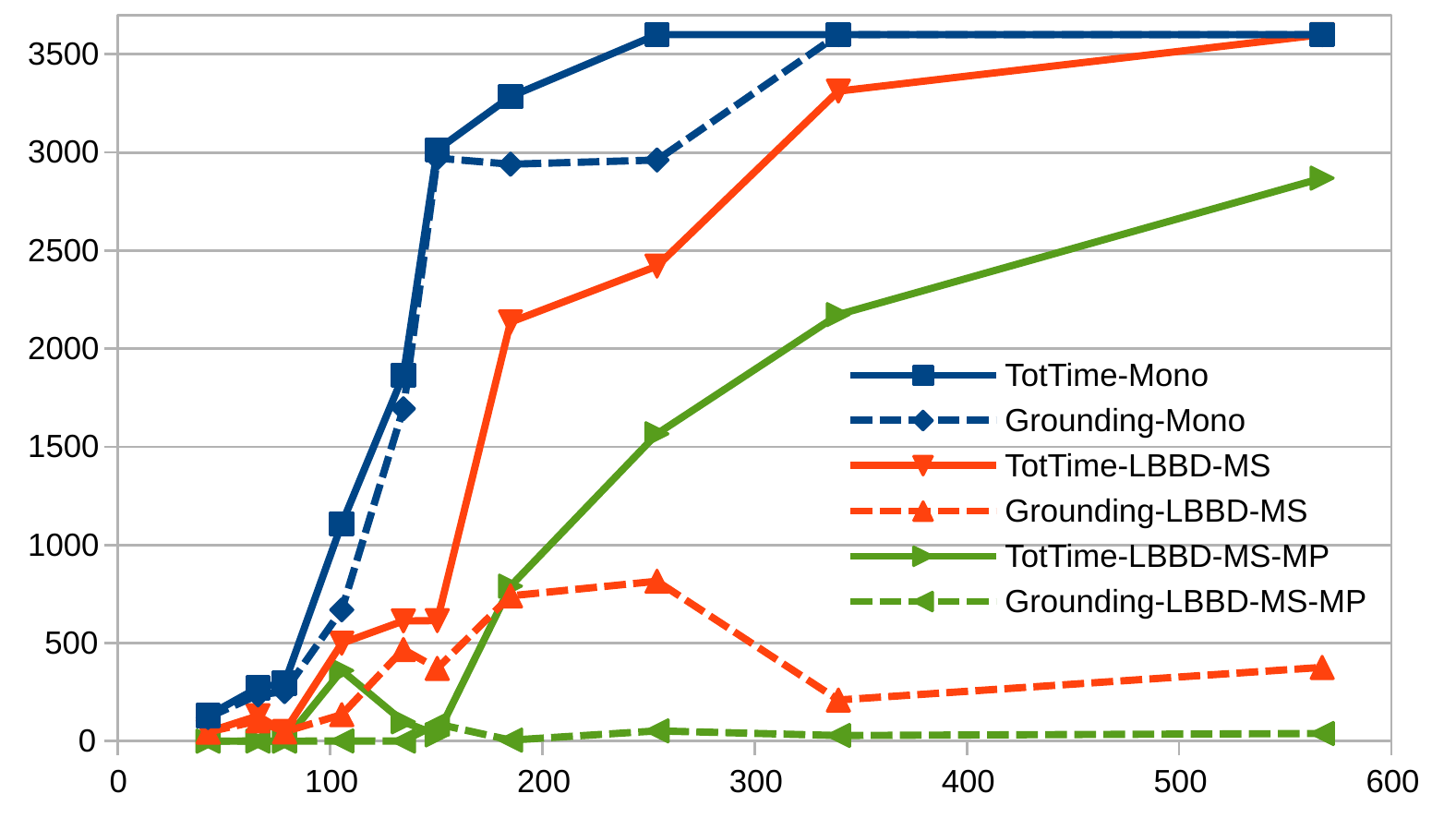}
    \caption{\label{fig:time_vs_serv} Run time vs number of services - all instances.}
\end{figure}

\begin{figure}
    \centering
    \includegraphics[width=\textwidth]{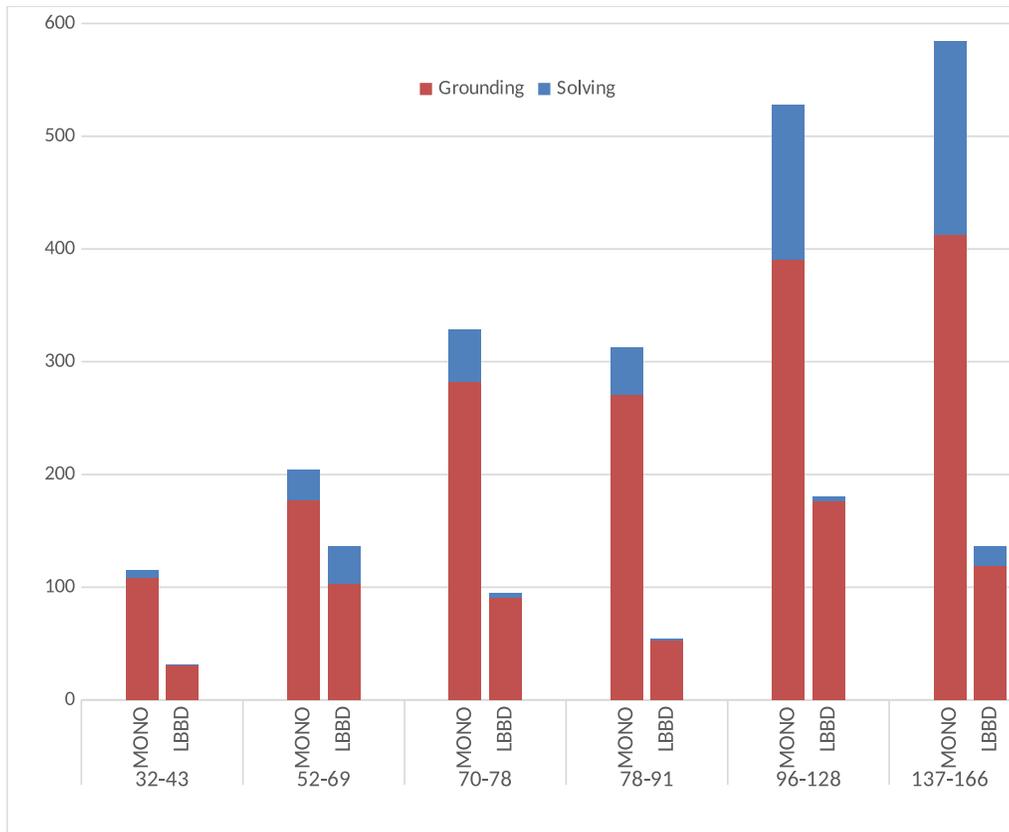}
    \caption{\label{fig:stacked} Run time  vs \# services instances without timeout}
\end{figure}

To have a finer detail on how  the running time varies with the size of the instance, we plot in Figure~\ref{fig:time_vs_serv} the runtime versus the number of services to be scheduled, comparing the monolithic approach and the LBBD method equipped with \ac{MS}. Solid lines represent total time, while dotted lines show the time required by grounding. Instances running into out of memory were counted as running for 3600s.
For the LBBD-MS we also show the time for the overall algorithm (orange series) as well as the time required by the  MP (green series). 
The time required by the SP can  be easily evaluated by difference between the orange and the green series.

We can observe the following facts: (i) as expected, the total time grows as the number of services increases regardless of the method used; (ii) for the monolithic approach, the running time is almost entirely spent in the grounding phase; (iii) for the LBBD with MS the grounding time spent in the MP is stable across instances and negligible; the grounding time of this approach is almost all due to the Sub-Problems.

To show the exact speedup in the two phases, Figure~\ref{fig:stacked} considers only those instances for which the monolithic approach was able to terminate within the timeout; clearly these are the most favorable for the monolithic approach. In these instances, the average grounding time for LBBD was 39.2\% of that of the monolithic, while the average solving time was 12.05\% of that of the monolithic, with almost an order of magnitude of improvement in the solving time.

\section{Conclusions}
\label{sec:conclusions}

\acresetall 

In this work, we adopted \ac{LBBD} in a solving process based on Answer Set Programming; to the best of our knowledge, this is the first work adopting this methodology with ASP.
\ac{LBBD} is a widely used solving technique in Operations Research and in Constraint Programming, and it constitutes one of the
most effective technologies for hybridization of Integer Linear Programming and Constraint Programming technologies.
With \ac{LBBD}, the problem can be decomposed  without losing completeness, i.e., maintaining the possibility to find the  optimal solution and prove its optimality.
Even more interestingly, this opens a new avenue of integration of ASP with other paradigms for solving constrained optimization problems, e.g. new hybrid algorithms involving  ASP  and Constraint Programming or Integer Linear Programming.

The considered application is a scheduling problem for chronic outpatients with Non-Communicable Diseases needing recurrent services at the hospital.
The experimental results show that \ac{LBBD} enlarges the applicability of ASP to larger 
instances without sacrificing optimality.
Future work 
includes strengthening the efficiency of the \ac{LBBD} scheme by providing stronger no-goods from the subproblems to the master problem, \added{and to apply LBBD to other problems.}

We believe that the \ac{LBBD} approach could be applied to
a number of applications already available for ASP (e.g., \cite{guido2020scheduling,cardellini2021two,CarusoJLC23},
just to name a few in the health care domain)
in which the global problem was greedily split into sub-parts;
we hope that this work could be of inspiration for the many ASP applications in which the authors forewent
obtaining optimality, and widen even further the ASP applications in the real world.

\bibliographystyle{acmtrans}
\bibliography{biblio}

\end{document}